\documentclass[12pt,a4paper,twoside]{article}
\makeatletter\if@twocolumn\PassOptionsToPackage{switch}{lineno}\else\fi\makeatother

\usepackage{amsfonts,amssymb,amsbsy,latexsym,amsmath,tabulary,graphicx,times,xcolor}
\usepackage[T1]{fontenc}
\usepackage{url,multirow,morefloats,floatflt,cancel,tfrupee,multicol}
\usepackage{comment}
\makeatletter

\AtBeginDocument{\@ifpackageloaded{textcomp}{}{\usepackage{textcomp}}}
\makeatother
\usepackage{colortbl}
\usepackage{xcolor}
\usepackage{pifont}
\usepackage[nointegrals]{wasysym}
\usepackage[superscript,biblabel]{cite}

\makeatletter

\@ifundefined{etal}{\def\etal{\textit{et~al}}}{}

\AtBeginDocument{
\expandafter\ifx\csname eqalign\endcsname\relax
\def\eqalign#1{\null\vcenter{\def\\{\cr}\openup\jot\m@th
  \ialign{\strut$\displaystyle{##}$\hfil&$\displaystyle{{}##}$\hfil
      \crcr#1\crcr}}\,}
\fi
}

\@ifundefined{tsGraphicsScaleX}{\gdef\tsGraphicsScaleX{1}}{}
\@ifundefined{tsGraphicsScaleY}{\gdef\tsGraphicsScaleY{.9}}{}
\def\checkGraphicsWidth{\ifdim\Gin@nat@width>\linewidth
	\tsGraphicsScaleX\linewidth\else\Gin@nat@width\fi}

\def\checkGraphicsHeight{\ifdim\Gin@nat@height>.9\textheight
	\tsGraphicsScaleY\textheight\else\Gin@nat@height\fi}

\def\fixFloatSize#1{}
\let\ts@includegraphics\includegraphics

\def\inlinegraphic[#1]#2{{\edef\@tempa{#1}\edef\baseline@shift{\ifx\@tempa\@empty0\else#1\fi}\edef\tempZ{\the\numexpr(\numexpr(\baseline@shift*\f@size/100))}\protect\raisebox{\tempZ pt}{\ts@includegraphics{#2}}}}

\AtBeginDocument{\def\includegraphics{\@ifnextchar[{\ts@includegraphics}{\ts@includegraphics[width=\checkGraphicsWidth,height=\checkGraphicsHeight,keepaspectratio]}}}

\DeclareMathAlphabet{\mathpzc}{OT1}{pzc}{m}{it}

\def\URL#1#2{\@ifundefined{href}{#2}{\href{#1}{#2}}}

\edef\fntEncoding{\f@encoding}

\makeatother

\newif\ifmultipleabstract\multipleabstractfalse%
\usepackage[hmargin=1.8cm,vmargin=2.2cm]{geometry}
\makeatletter
\def\author#1{\gdef\@author{\hskip-\dimexpr(\tabcolsep)\hskip1pt\parbox{\dimexpr\textwidth-1pt}{\centering #1}}}

\let\@articletype\@empty \def\articletype#1{\gdef\@articletype{{\fontsize{14}{16}\selectfont #1}}}

\usepackage{fancyhdr}

\fancypagestyle{headings}{\fancyhf{}\fancyhead[RE]{\MakeTextUppercase{\textbf{\RunningAuthor}}}\fancyhead[LE]{\textbf{\thepage}}\fancyhead[LO]{\MakeTextUppercase{\textbf{\RunningHead}}}\fancyhead[RO]{\textbf{\thepage}}}
\fancypagestyle{plain}{\fancyhf{}\fancyfoot[C]{\textbf{\thepage}}}

\AtBeginDocument{\usepackage{lastpage}}
\def\title#1{%
  \gdef\@title{%
    \ifx\@articletype\@empty\else\@articletype~\\\fi%
     #1}%
}

\usepackage{abstract}
\def\abstractname{\textbf{Abstract}}
\renewenvironment{onecolabstract}
{\vspace*{-.4pc}\trivlist\item[]\leftskip1pt\noindent\selectfont\hfill\abstractname\hfill\mbox{\null}\par\ignorespaces}{\endtrivlist}

\def\NormalBaseline{\def\baselinestretch{1.1}}

\usepackage{textcase}
\usepackage[explicit]{titlesec}
\titleformat{\section}[block]{\NormalBaseline\boldmath\bfseries}
{\thesection.}
{6pt}
{#1}
[]
\titleformat{\subsection}[hang]{\NormalBaseline\filright\itshape}
{\thesubsection.}
{6pt}
{#1}
[]
\titleformat{\subsubsection}[runin]{\NormalBaseline\filright\itshape}
{\hspace{16pt}\thesubsubsection}
{6pt}
{#1}
[]
\titleformat{\paragraph}[runin]{\NormalBaseline}
{\theparagraph}
{6pt}
{#1}
[]
\titleformat{\subparagraph}[runin]{\NormalBaseline}
{\thesubparagraph}
{6pt}
{#1}
[]

\titlespacing{\section}{0pt}{1.5\baselineskip}{.2\baselineskip}  
\titlespacing{\subsection}{0pt}{1.5\baselineskip}{.2\baselineskip}  
\titlespacing{\subsubsection}{0pt}{1.5\baselineskip}{.2\baselineskip}  
\titlespacing{\paragraph}{0pt}{.5\baselineskip}{10pt}  
\titlespacing{\subparagraph}{0pt}{.5\baselineskip}{10pt}

\usepackage{caption}
\DeclareCaptionLabelFormat{figlabel}{Figure #2}
\date{}
\makeatother

\usepackage{float, makecell}
\usepackage{xr}
\usepackage{pdfpages}
\renewcommand{\thesubsection}{\Alph{subsection}}
\renewcommand{\thesubsubsection}{\Roman{subsubsection}}
\usepackage{bm}

\begin{document}
\title{SPADA: A Toolbox of Designing Soft Pneumatic Actuators for Shape Matching based on Surrogate Modeling}
\def\RunningHead{SPADA}
\def\RunningAuthor{YAO \etal}

\author{Yao Yao$^{1}$, Liang He$^{2}$ and Perla Maiolino$^{1,3}$ 
\thanks{$^{1}$ Yao Yao and Perla Maiolino are with Oxford Robotics Institute, University of Oxford, Oxford, OX1 2JD, United Kingdom; $^{2}$ Liang He is with Institute of Biomedical Engineering, University of Oxford, Oxford, OX1 2JD, United Kingdom; $^{3}$ Perla Maiolino is also with the University of Genoa, Liguria, IT. Corresponding Author: Yao Yao. 
{\tt\small yao.yao/liang.he/perla.maiolino@eng.ox.ac.uk}}%
}

\maketitle


{\begin{onecolabstract}

Soft pneumatic actuators (SPAs) produce motions for soft robots with simple pressure input, however they require to be appropriately designed to fit the target application.  Available design methods employ kinematic models and optimization to estimate the actuator response and the optimal design parameters, to achieve a target actuator’s shape. Within SPAs, Bellow-SPAs excel in rapid prototyping and large deformation, yet their kinematic models often lack accuracy due to the geometry complexity and the material nonlinearity.  Furthermore, existing shape-matching algorithms are not providing an end-to-end solution from the desired shape to the actuator. In addition, despite the availability of computational design pipelines, an accessible and user-friendly toolbox for direct application remains elusive. This paper addresses these challenges, offering an end-to-end shape-matching design framework for bellow-SPAs to streamline the design process, and the open-source toolbox SPADA (\textbf{S}oft \textbf{P}neumatic \textbf{A}ctuator \textbf{D}esign fr\textbf{A}mework) implementing the framework with a GUI for easy access. It provides a kinematic model grounded on a modular design to improve accuracy, Finite Element Method (FEM) simulations, and piecewise constant curvature (PCC) approximation. An Artificial Neural Network-trained surrogate model, based on FEM simulation data, is trained for fast computation in optimization. A shape-matching algorithm, merging 3D PCC segmentation and a surrogate model-based genetic algorithm, identifies optimal actuator design parameters for desired shapes. The toolbox, implementing the proposed design framework, has proven its end-to-end capability in designing actuators to precisely match 2D shapes with root-mean-square errors of 4.16, 2.70, and 2.51mm, and demonstrating its potential by designing a 3D deformable actuator.

\def\keywordstitle{Keywords}
\end{onecolabstract}}
 
\begin{multicols}{2}

\section{Introduction}
Soft pneumatic actuators (SPAs) have a significant impact on the soft robotics field by enabling a variety of applications of soft robotics, such as grasping~\cite{grasping, grasping2}, locomotion~\cite{locomotion, locomotion2}, and manipulation~\cite{manipulation, manipulation2}. 
These versatile functionalities come from not only the inherent flexibility and adaptability of soft materials but also the capacity of SPAs to produce diverse 2D and 3D motion patterns~\cite{SPA,SPA2,SPA3}. This is achieved by pre-programming the morphology of the actuator to achieve complex motions from a single pressure input thus reducing the need of complex control~\cite{programmable, programmable2, programmable3, programmable4}. This trait has been widely employed in designing soft robots, especially those that draw inspiration from biological entities, as the desired actuator shape can be derived from creatures' movement.  For instance, the locomotion pattern of inchworms has been replicated by designing actuators to recursively crawl~\cite{locomotion2}, while the grasping profile of an elephant trunk has been emulated by actuators with helical configurations to achieve versatile and secure holds~\cite{manipulation2}.

Despite considerable advancements in discovering suitable soft actuator structures and materials to mimic biological motions, the design process is impeded by complexities tied to the dynamic, contact, and multimodal modeling of these actuators~\cite{SPA2, SPA3}. Addressing these challenges remains difficult, given the limitation of current modeling and computational approaches in managing large geometric deformation and material nonlinearity~\cite{modeling}. As a result, the research community's current primary focus lies in designing soft actuators for specific profiles or trajectories with kinematic models and optimization methods~\cite{Jiang2021}by establishing systematic methodologies and strategies to enhance efficiency~\cite{Connolly2017, Singh2020, Jiang2021, programmable3}.

Fiber-reinforced SPAs, recognized for their versatile applications~\cite{FR, locomotion2, FR2}, leverage fiber arrangements for varied 2D and 3D motion~\cite{programmable}. Connolly et al.\cite{Connolly2017} presented an analytical model for these SPAs, comprising distinct modules for bending, twisting, elongating, and expanding. They took a predefined kinematic trajectory for each module and applied optimization to identify optimal actuator design parameters for matching shapes along the trajectory. Singh and Krishnan~\cite{Singh2020}, concentrating on bending modules, proposed a method that segments 2D curves into piecewise constant curvature (PCC) sections, with design parameters derived from their analytical model. However, the complex manufacturing of fiber-reinforced SPAs, especially the fiber routing process, faces challenges such as extended production time, inconsistency issues, and design limitations.

PneuNets actuators, formed of interconnected, pleated channels within an elastomer, deform when pressurized~\cite{PneuNet}. Their behavior is defined by altering channel geometry or material distribution~\cite{programmable2}. Advances in 3D printing allow these actuators to be crafted either via 3D-printed molds and silicon casting~\cite{Jiang2021}or direct 3D printing with flexible materials~\cite{PneuNet3DPrint_ElephantTrunk}. Following Connolly et al.'s strategy~\cite{Connolly2017}, Jiang et al.~\cite{Jiang2021} modeled PneuNets, which include bending, twisting, and helical modules. Their design process starts with manual 3D curve segmentation, followed by their analytical model and optimization to determine the actuator design for the target shape.

Transitioning to bellow-SPAs, replacing PneuNets' sharp channels with bellow-shaped convolutions, leads to an unfolding structure under pressure~\cite{Bellow, Bellow2}. This feature enables transferring material strain into structural deformation, making bellow-SPAs particularly suitable for 3D printing materials like Agilus30\texttrademark due to their lower elongation-at-break compared to silicon elastomers ~\cite{StratasysAgilus30}, ensuring precision in production. Kan et al.~\cite{programmable3} proposed an analytical model for modularized bellow-SPA designs, using interconnected channels to achieve varied deformation curves. They employed a sampling-based optimization to design channels for a desired end-tip trajectory.

\begin{figure*}
    \centering
    \includegraphics[width=0.9\textwidth]{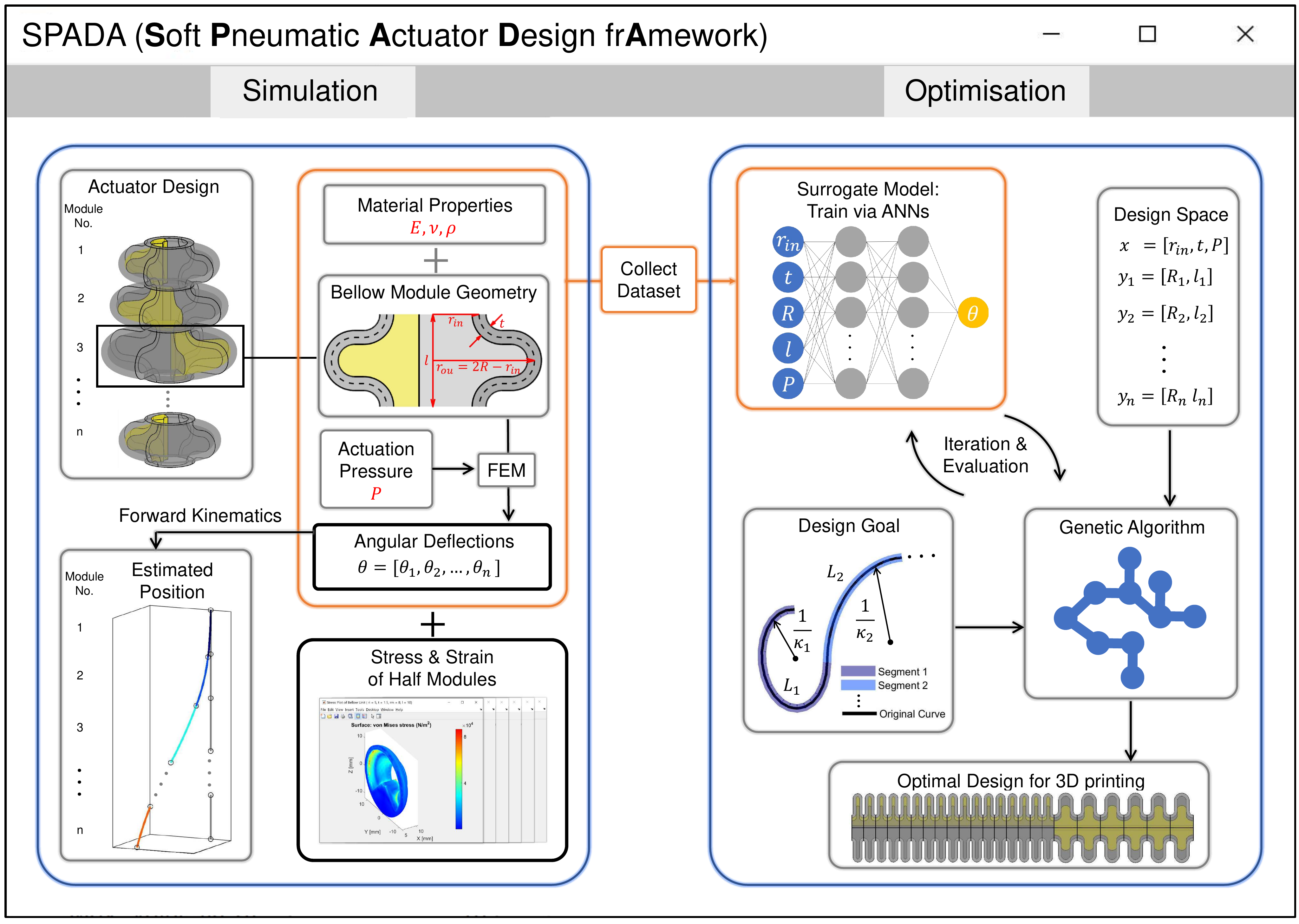}
    \caption{An overview of the bellow soft pneumatic actuators (SPAs) design toolbox, consisting of two main components: the simulation and the optimization components. The simulation component (left) facilitates the design of bellow SPAs by allowing modification of material properties, such as Young's Modulus, Poisson's ratio and density, adjusting the modular geometric parameters, and customizing module assemblies. It utilizes the Finite Element Method (FEM) to simulate each module's angular deflection, mechanical stress, and strain based on the defined actuation pressure and predicts the actuator's configuration through forward kinematics. It also collects a dataset of the FEM model. The optimization component (right) uses an Artificial Neural Network (ANN) to train on the dataset to provide a surrogate model for optimization. It takes an input of a target shape and divides it into constant curvature segments with length and curvature parameters using a 3D piecewise constant curvature segmentation. Then, a genetic algorithm is employed to optimize the parameters in the design space of a bellow SPA. Finally, a CAD file of the bellow SPA based on the optimal design parameter for matching the target shape can be generated.}
    \label{fig:schematic}
\end{figure*}

Analytical models, although widely used for forward kinematics, often struggle to accurately predict substantial deformations due to their reliance on geometric simplifications coupled with material nonlinearity. The Finite Element Method (FEM) is known for its accurate predictions~\cite{FEM} but is computationally expensive, making it unsuited for repeated optimization~\cite{optimization, optimization2}. Recent advances use machine learning, particularly supervised artificial neural networks (ANNs) trained with FEM data, to create efficient surrogate models for bellow-SPAs~\cite{Ali2021}. However, these models have covered limited actuator design spaces~\cite{surrogate}. 

Furthermore, current shape-matching optimization methods for SPAs are limited by the need for human intervention in segmenting 3D shapes and in converting optimal parameters into a manufacturable design. This results in a lack of an end-to-end solution for seamlessly connecting the desired shape to a ready-to-print actuator design file. Furthermore, the inherent complexities of implementing design frameworks emphasize the need for an open-source, user-friendly design toolbox to enhance accessibility and efficiency~\cite{pushbutton}. Presently, the available toolboxes for bellow-SPAs exhibit some limitations, either focusing solely on simulation or covering a limited design space~\cite{Yao2022,toolbox2,toolbox3}. 
On the other hand, some permit optimization but lack a specific focus on shape-matching~\cite{toolbox3,toolbox4}.

Therefore, to address the above challenges, this article provides an end-to-end shape-matching design framework for bellow-SPAs, and an open-source toolbox named SPADA (\textbf{S}oft \textbf{P}neumatic \textbf{A}ctuator \textbf{D}esign fr\textbf{A}mework) that implements this framework through a user-friendly GUI. Fig.~\ref{fig:schematic} shows the schematic of the functions included in this design toolbox, consisting of simulation and optimization functionalities. The simulation component permits modular design customization, defining material properties, and predicting actuator kinematics using FEM combined with PCC approximation. It also allows for collecting a FEM dataset based on varied geometric parameters and actuation pressures. The optimization component can generate a surrogate model, trained on FEM or self-characterized data, to expedite computations.  It segments input 2D/3D curves using PCC segmentation and uses the surrogate model to optimize the bellow-SPA's design parameters for shape-matching.

The rest of this article is organized as follows: Section 2 details the shape-matching design method, covering the bellow-SPA's kinematics, FEM simulation, surrogate modeling, and the shape-matching optimization for determining design parameters. Section 3 introduces the SPADA toolbox, implementing the discussed framework with a user-friendly GUI for kinematic analysis and shape-matching optimization. Section 4 applies the toolbox for precise 2D shape-matching designs, also exploring 3D actuator design potential.

\section{DESIGN AND METHOD}
\label{sec:designandmethod}

\begin{figure*}
    \centering
    \includegraphics[width=0.9\textwidth]{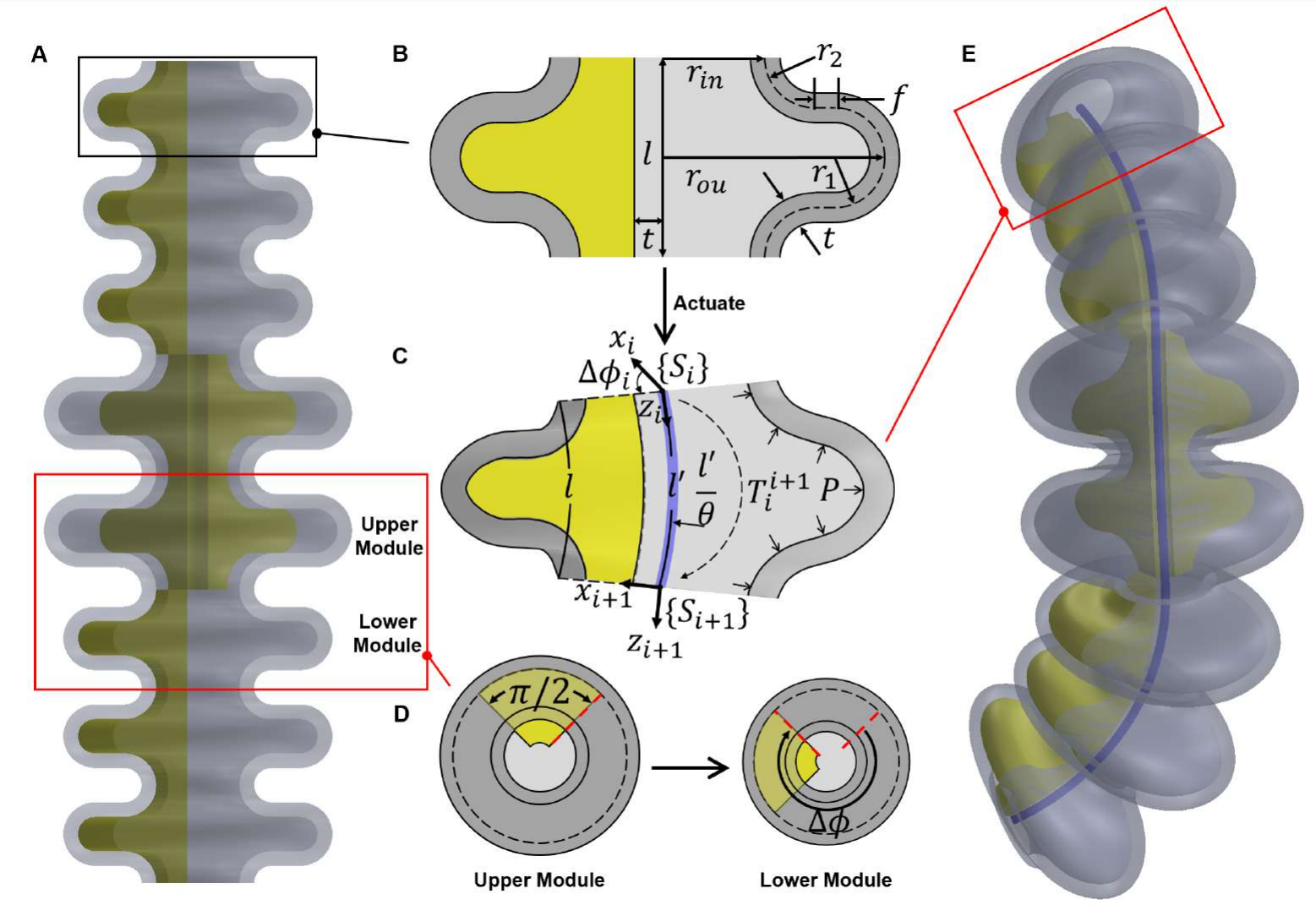}
    \caption{The design and kinematics of the bellow soft pneumatic actuators (SPAs). A) A bellow-SPA consists of modules. B) The parameterized 2D geometry of the module. C) The deformed module and its kinematics model represented by an arc with length and curvature. D) Rotation angle between adjacent modules. E) The deformed bellow-SPA upon pressurization with an inner line representing its deformed shape.}
    \label{fig:design}
\end{figure*}

This section describes the methodologies employed for designing, analyzing, and optimizing bellow-SPAs for shape-matching. We present a modular design, describe its kinematics, discuss FEM simulations and ANN training and introduce a shape-matching optimization algorithm for desired actuator shapes.

\subsection{bellow-SPAs Design and Kinematics}
\label{sec:design}

Modularity has demonstrated its effectiveness in exploring large design space of SPAs for achieving desired behaviors~\cite{programmable, programmable2, programmable3, programmable4}. This approach has been uniformly adopted in prior design methodologies for shape-matching, owing to its capability to accelerate kinematics prediction and optimization.

As shown in Fig.~\ref{fig:design}, a bellow-SPA (Fig.~\ref{fig:design}A) can be built by stacking modules, whose deformation is determined by the deformation of each module along with the rotation angle between two adjacent modules. Each module consists of a U-shaped bellow ring and a $\frac{\pi}{2}$ fan-shaped constraint (Fig.~\ref{fig:design}B), parameterized by the geometric parameters in Table~\ref{tab:geometryparameter}.

\begin{center}
    \captionof{table}{Geometric Parameters of the bellow-SPA Module.}
    \footnotesize
    \begin{tabular}{c c c}
        \hline
        Parameters & \makecell{Symbol/\\Equation} & \makecell{Default Value\\ (mm)}\\
        \hline
        Module Length & $l$ & 10 \\
        \hline
        Inner Radius & $r_{in}$ & 5\\
        \hline
        Average Radius & $R$ & 8\\
        \hline
        Outer Radius & $r_{ou} = 2R - r_{in}$ & 11\\
        \hline
        Upper Radius & $r_1 = l/4$ & 2.5\\
        \hline
        Lower Radius & $r_2 = l/4$ & 2.5\\
        \hline
        Wall Thickness & $t$ & 1.5\\
        \hline
        Flank Length & $f = r_{ou} - r_{in} - r_1 - r_2$ & 1\\
        \hline
    \end{tabular}
    \label{tab:geometryparameter}
\end{center}

Essentially, the geometry of the bellow-SPA module is governed by four parameters $r_{in}$, $t$, $R$ and $l$, bounded by their non-negativity,

\begin{equation}
    \begin{cases}
        l > 2t\\
        R - r_{in} \geq \frac{l}{4}\\
    \end{cases}
\end{equation}

The constraint restricts the expansion of the module, allowing it to unfold only in opposite directions during pressurization, resulting in a bending deformation of angle $\theta$ under the applied pressure $P$.
Assuming the deformed module remains its original length along an axis on the side of constraint, offset by a distance of $r_{in}+\frac{t}{2}$ from the central axis, and the deformation follows the CC approximation, it can be represented by an associate arc (indicated by a line) with the length $l'$ and curvature $\frac{\theta}{l'}$ ( Fig.~\ref{fig:design}C), where $l' = \theta (\frac{l}{\theta} + r_{in} + \frac{t}{2})$.

As shown in Fig.~\ref{fig:design}D, by having a clockwise rotation angle $\Delta \phi$ between two adjacent modules (from upper to lower), the designed actuator can achieve spatial deformation, where the inner line indicates its deformed shape (Fig.~\ref{fig:design}E).
To further simplify the design, all stacking modules share the same inner radius $r_{in}$, wall thickness $t$, actuation pressure $P$ and the manufacturing material. To this end, the kinematics model of bellow-SPAs can be derived using the transformation matrix $T^{i+1}_i$ mapping from the arc base frame $\{S_i\}$ to the tip frame $\{S_{i+1}\}$ of the $i$-th module, where
\begin{equation}
\scriptsize
\begin{aligned}
&T^{i+1}_i = \\
&\begin{bmatrix}
c{\Delta \phi_i}c{\theta_i} & -s{\Delta \phi_i} & c{\Delta \phi_i}s{\theta_i}  & \frac{l'_i}{\theta_i}c{\Delta \phi_i}(1-c{\theta_i})\\
s{\Delta \phi_i}c{\theta_i} & c{\Delta \phi_i} & s{\Delta \phi_i}s{\theta_i}  & \frac{l'_i}{\theta_i}s{\Delta \phi_i}(1-c{\theta_i})\\
-s{\theta_i} & 0 & c{\theta_i} & \frac{l'_i}{\theta_i}s{\theta_i}\\
0 & 0 & 0 & 1
\end{bmatrix}
\end{aligned}
\normalsize
\end{equation}
\noindent
, in which $c$ and $s$ represent $\cos$ and $\sin$.

\begin{figure*}
    \centering
    \includegraphics[width=0.95\textwidth]{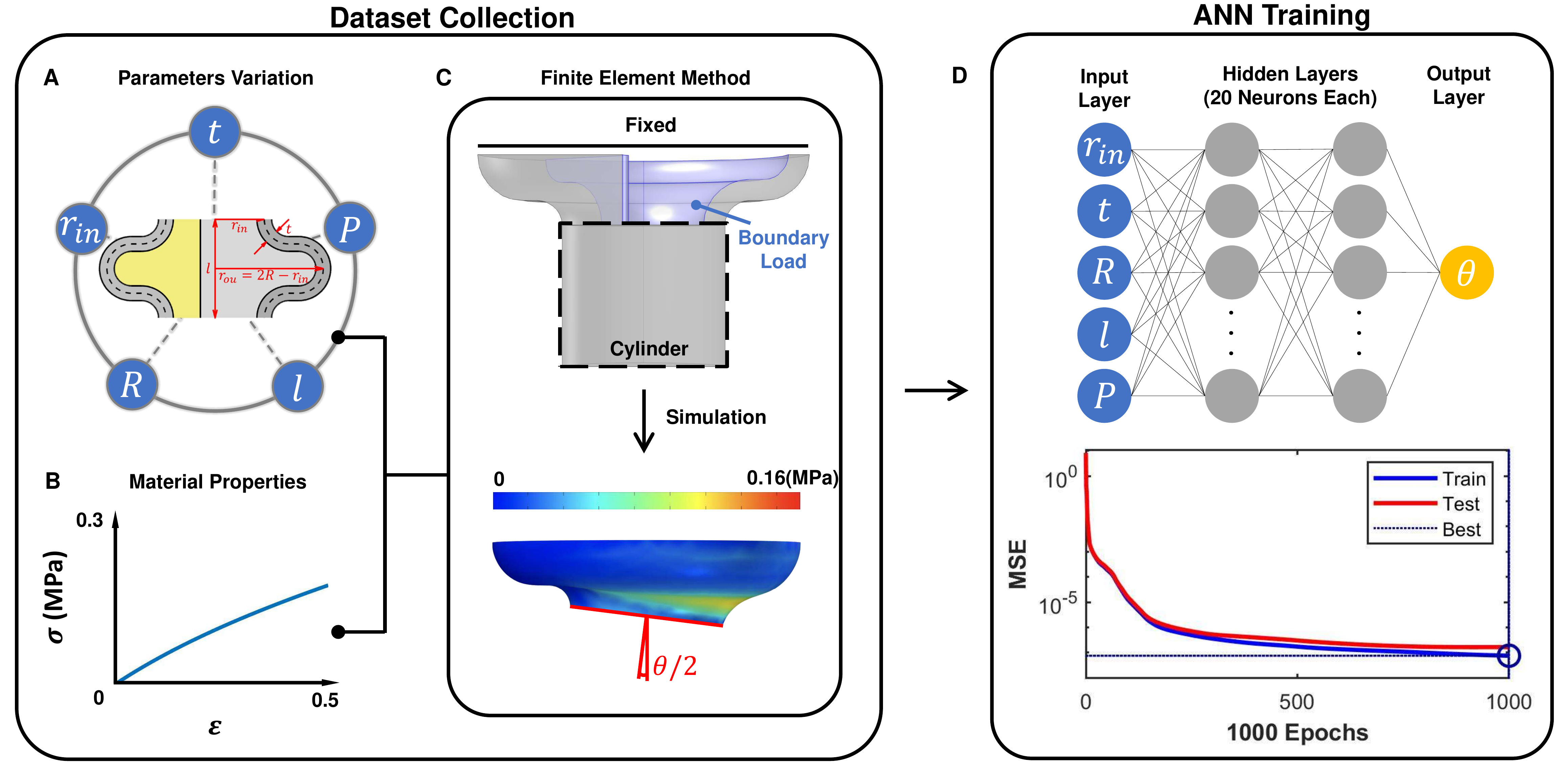}
    \caption{The process of generating the surrogate model. A) The design parameters that are varied in dataset collection. B) The elastic properties of the material. C) Boundary conditions of the half bellow module in the Finite Element Method (FEM) simulation from the cross-section perspective (top) and the deformed stress plot after simulation (bottom). D) The Artificial Neural Network (ANN) that is used to train the surrogate model based on the dataset (top), and the mean square error (MSE) of both training and test data in relation to the change of epochs (bottom).}
    \label{fig:surrogate}
\end{figure*}

\subsection{Surrogate Model based on the FEM dataset and ANN training}
The above kinematics model for bellow-SPAs maps arc parameters to actuator position but doesn't define the transformation from design space to arc parameters. Hence, a model connecting the design parameters (geometry and material properties) of a module to its deformation is needed. Fig.~\ref{fig:surrogate} illustrates the surrogate modeling process using a FEM model for dataset generation, subsequently training an ANN to expedite computation for optimization.

\subsubsection{FEM Simulation and Dataset Collection ~ ~}
\label{sec:FEM}

To generate a viable surrogate model of the bellow-SPA, a dataset should be obtained by sampling its feasible feature space. Four attribute geometry parameters and the actuation pressure are chosen as input to obtain a dataset (Fig.~\ref{fig:surrogate}). The ranges and intervals of the selected input are set as shown Table~\ref{tab:dataset}.

\begin{center}
    \captionof{table}{Range and interval of FEM simulation input data for dataset collection.}
    \footnotesize
    \begin{tabular}{c c c c}
        \hline
        Parameters & Minimum & Maximum & Interval\\
        \hline
        $r_{in}$ (mm) & 2 & 10 & 2 \\
        \hline
        $t$ (mm)& $r_{in}/4$ & $r_{in}/3$ & $r_{in}/24$\\
        \hline
        $R$ (mm)& $r_{in}+t$ & $2r_{in}$ & $(r_{in}-t)/4$\\
        \hline
        $l$ (mm)& $4t$ & $4(R - r_{in})$ & $R - r_{in} - t$\\
        \hline
        $P$ (kPa)& 0 & 10 & 0.5\\
        \hline
    \end{tabular}
    \label{tab:dataset}
\end{center}
Considering the limitation of the material customizability, the dataset is material specified. Therefore, a new set of collection is required for a new material. The incompressible Neo-Hookean hyperelastic material model is applied to the material (Fig.~\ref{fig:surrogate}B). 
 
\begin{equation}
\label{eq:neohookean}
    \begin{aligned}
    W = C_1(I_1 - 3) = \frac{\mu}{2}(I_1 - 3)
    \end{aligned}
\end{equation}
where $\mu = E/2(1+\nu)$ is the shear modulus, $I_1$ is the first invariant of the right Cauchy-Green deformation tensor, $E$ is Young's modulus and $\nu$ is the Poisson's ratio. Agilus30\texttrademark ~is used as the default material as it is a commonly used material for designing soft robots~\cite{Agilus30softrobot}. Material properties are obtained from a characterization work~\cite{Agilus30} and an official datasheet~\cite{StratasysAgilus30}.

A MATLAB\textsuperscript{\textregistered} script is written to automatically create FEM simulations in COMSOL MultiPhysics\textsuperscript{\textregistered} based on the input, and extract simulation results as output. Thanks to the symmetry of the bellow geometry, the FEM simulation is performed on the half bellow module as a stationary study of the 3D solid mechanics model (top of Fig.~\ref{fig:surrogate}C). The top end is fixed, and the other is connected to a cylinder to smooth the element deformation. A uniformly distributed load is applied to the surface colored in blue.

From the nonlinear simulation, deformation of the half bellow module and its von Mises stress distribution can be obtained (bottom of Fig.~\ref{fig:surrogate}C). By using the displacement of a group of points at the end of the cylinder, the angular deflection $\theta$ of the bellow module can be calculated (see details of the FEM and material model in Supplementary Data A). 

\subsubsection{Surrogate Model Trained via an ANN ~ ~ ~}
\label{sec:ANN}
To generate the surrogate model, a feedforward ANN (top of Fig.~\ref{fig:surrogate}D) was constructed with an input layer of five neurons for the input data (consisting of four geometric parameters and the actuation pressure based on Table~\ref{tab:dataset}, two hidden layers of twenty neurons each and an output layer of one neuron for the corresponding output data (angular deflection). The mean squared error (MSE) served as the performance function. The Bayesian Regularization backpropagation is used as the training function.

The pre-collected FEM simulation dataset is randomly divided into training and test sets (80\% and 20\%, respectively) and trained until 1000 epochs have been reached. After training, the MSE reaches below $10^{-6}$ for both training and test data (bottom of Fig.~\ref{fig:surrogate}D). This pre-trained network will be then used as the surrogate model in the form of $\theta = f(r_{in}, t, R, l, P)$ to quickly predict the angular deflection of the bellow-SPA module for a given set of design parameters, allowing for efficient design exploration and optimization.

\subsection{Shape-Matching Optimization}
\label{sec:opt}
\begin{figure*}
    \centering
    \includegraphics[width=0.9\textwidth]{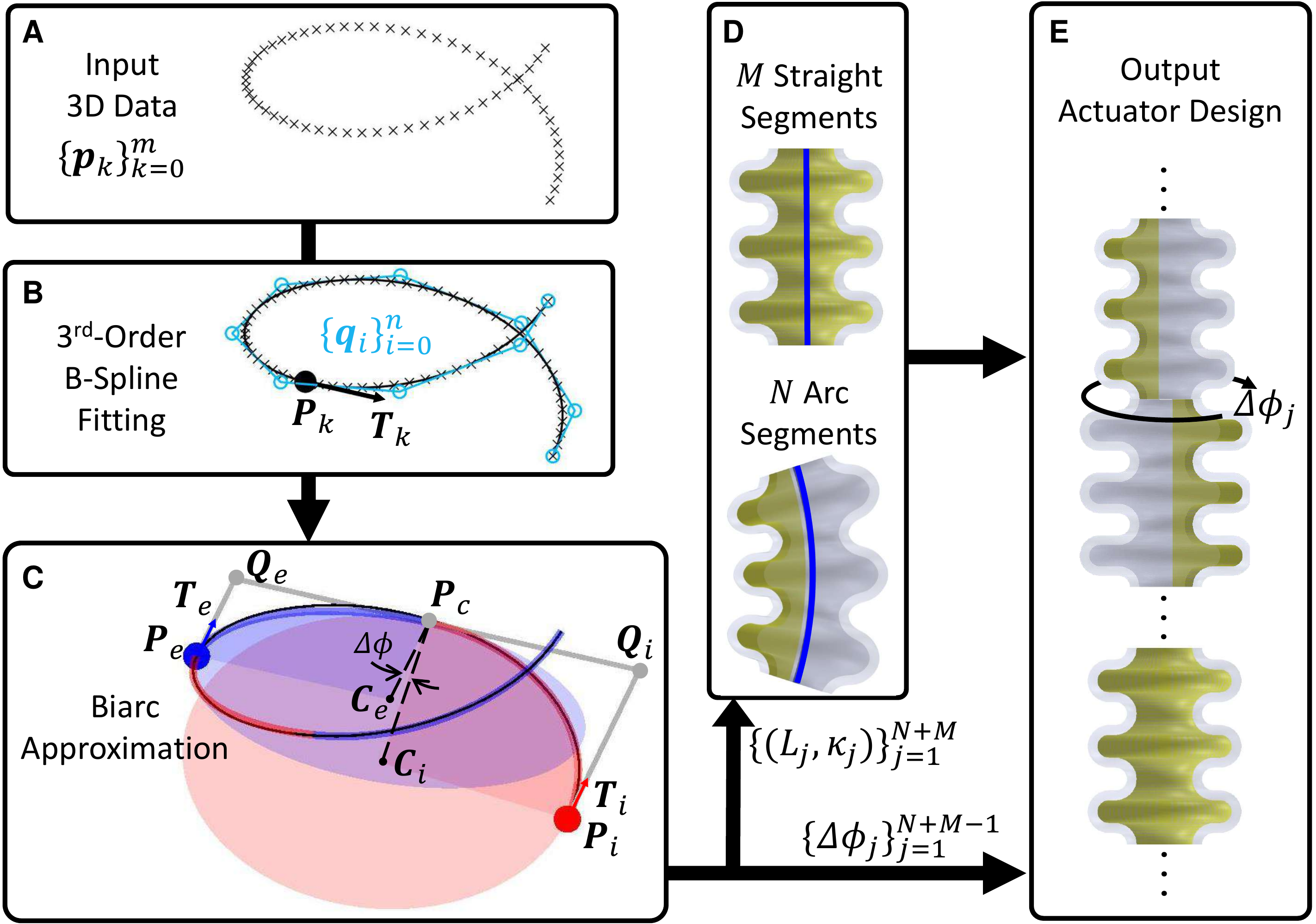}
    \caption{The shape-matching optimization process, explained in Sec.~\ref{sec:opt} in detail. A)-C) shows the piecewise constant curvature segmentation algorithm, in which the input data $\{\bm{p}_k\}^m_{k=0}$ is first fitted by a 3rd-order B-spline curve controlled by light blue control points $\{\bm{q}_i\}^n_{i=0}$, with projected point $\bm{P}_k$ and its tangent vector $\bm{T}_k$; then discretized into constant curvature segments using the biarc approximation by interpolating between two end points $\bm{P}_i$, $\bm{P}_e$ and their tangent vectors $\bm{T}_i$, $\bm{T}_e$, based on the control points $\bm{Q}_i$, $\bm{Q}_e$ and the connection point $\bm{P}_c$; the rotation angle $\Delta \phi$ between two segments is $\angle \bm{C}_i \bm{P}_c \bm{C}_e$. D) shows that for each CC segment represented by shape parameters ($L_j$, $\kappa_j$), a module will be designed to match its shape by stacking one or more identical modules. E) shows the output actuator is designed by stacking the previous modules in order with rotation angles $\Delta \phi_j$ between segments.
    }
    \label{fig:shapematching}
\end{figure*}

Given a curve represented by a set of 3D ordered points $\{\bm{p}_k\}^m_{k=0}$, we aim to design a bellow-SPA to deform from an unactuated straight line to a desired shape upon pressurization. Fig.~\ref{fig:shapematching} outlines our shape-matching process. Initially, A 3D PCC segmentation algorithm discretizes the curve into CC segments (Fig.~\ref{fig:shapematching}A-C). Next, the actuator's design involves determining the optimal module parameters to match each CC segment (Fig.~\ref{fig:shapematching}D-E). Notably, "segment" pertains to the desired shape's divided part, while "module" refers to the bellow-SPA's building block. Each segment's shape is approximated using one or multiple identical modules.

PCC requires the curve to follow $G^1$ continuity, meaning two adjacent CC segments share an endpoint with aligned tangent vectors. The biarc method, frequently used for this~\cite{biarc}, requires two endpoints $\bm{P}_i$, $\bm{P}_e$ and their tangents $\bm{T}_i$, $\bm{T}_e$ (Fig.~\ref{fig:shapematching}C). The challenge is determining the connection point $\bm{P}_c$ between two CC segments.

\begin{equation}
\bm{P}_c = \frac{d_2(\bm{P}_i + d_1 \bm{T}_i) + d_1(\bm{P}_e - d_2 \bm{T}_e)}{d_1 + d_2}
\end{equation}
where $d_1 = |\bm{Q}_i - \bm{P}_i| = |\bm{Q}_i - \bm{P}_c|$ and $d_2 = |\bm{Q}_e - \bm{P}_e| = |\bm{Q}_e - \bm{P}_c|$, with $\bm{Q}_i$ and $\bm{Q}_e$ as biarc control points. The relationship $|\bm{Q}_e - \bm{Q}_i| = d_1 + d_2$ allows solving the biarc with $d_1 = d_2$. Furthermore, the rotation angle $\Delta \phi$ between two CC segments is determined by the angle between $\bm{C}_i\bm{P}_c$ and $\bm{C}_e\bm{P}_c$, where $\bm{C}_i$ and $\bm{C}_e$ are the respective circle centers. If a segment is straight, it has no rotation with adjacent segments.

The biarc approximation samples data points, starting from the initial point, seeking the longest biarc that approximates data within a preset tolerance. This process is iteratively repeated until all data are approximated~\cite{PCC}. Tangent vectors are derived using a 3rd-order B-spline curve fitted to the original 3D data (Fig.~\ref{fig:shapematching}B), enabling the calculation of projection points and their tangent vectors. This B-spline curve is defined by $n+1$ control points $\{\bm{q}_i\}^n_{i = 0}$,

\begin{equation}
\bm{S}(t) = \sum^n_{i=0} B_{i,d}(t)\bm{q}_i
\end{equation}

where $d = 3 - 1$ is the degree of curve, 3 is the order of curve, $t$ is a sequence of non-decreasing uniform knots that $t_i = \frac{i-d}{n+1-d}$ for $d+1 \leq i \leq n$, and there is $d+1$ number of $0$ and $1$ at the beginning and end of the knot sequence, respectively. $B$ is the basis function that is defined recursively,
\begin{equation}
    B_{i,0}(t) = 
    \begin{cases}
        1, \ t_i \leq t < t_{i+1}\\
        0, \ \textrm{otherwise}
    \end{cases}
\end{equation}

\begin{equation}
    \begin{aligned}
        B_{i,j}(t) & = \frac{t - t_i}{t_{i+j}-t_i} B_{i,j-1}(t)\\
        & + \frac{t_{i+j+1} - t}{t_{i+j+1}-t_{i+1}} B_{i+1,j-1}(t)
    \end{aligned}
\end{equation}

Assuming that the sample sequence of the input data points is $t_k = \frac{k}{m}$, the matrix $\hat{q}$, composed of the control points of the B-spline that fits the data can be obtained using the least-squares fitting,
\begin{equation}
    \mathbf{\hat{q}} = ((\mathbf{A}^T \mathbf{A})^{-1} \mathbf{A}^T)\mathbf{\hat{p}}
\end{equation}
where $\mathbf{A} = [B_{i,d}(t_k)]$ and $\mathbf{\hat{p}}$ is the matrix composed of all data points, and the number of control points $n+1$ can be determined by minimising the error from input data points to the B-spline curve~\cite{b-spline, matlabB-spline}. Then, all data points are projected to the B-spline curve to get the projected points $\{\bm{P}_k\}^m_{k=0}$ and their tangent vectors $\{\bm{T}_k\}^m_{k=0}$.

A number of CC segments are obtained after the PCC segmentation (Fig.~\ref{fig:shapematching}D). If a segment curvature is below $10^{-3}$, These segments are designed by stacking modules filled with internal constraints. For arc segments, a genetic algorithm determines the bellow-SPA module design parameters for matching shapes by stacking identical modules. The selection of the genetic algorithm is informed by our previous work\cite{Yao2023} thanks to its compatibility with surrogate modeling and its capacity to effectively handle a large number of design variables. The objective is to minimize the difference in arc parameters (arc length and curvature) between the deformed segment shape (composed of designed modules and predicted by a surrogate model) and the desired segment shape (see details in Supplementary Data B). The final actuator is constructed by orderly stacking these modules, considering rotation angles between segments (Fig.~\ref{fig:shapematching}E).

\section{The Toolbox Implementation}
\label{sec:toolbox}

\begin{figure*}
    \centering
    \includegraphics[width=0.99\textwidth]{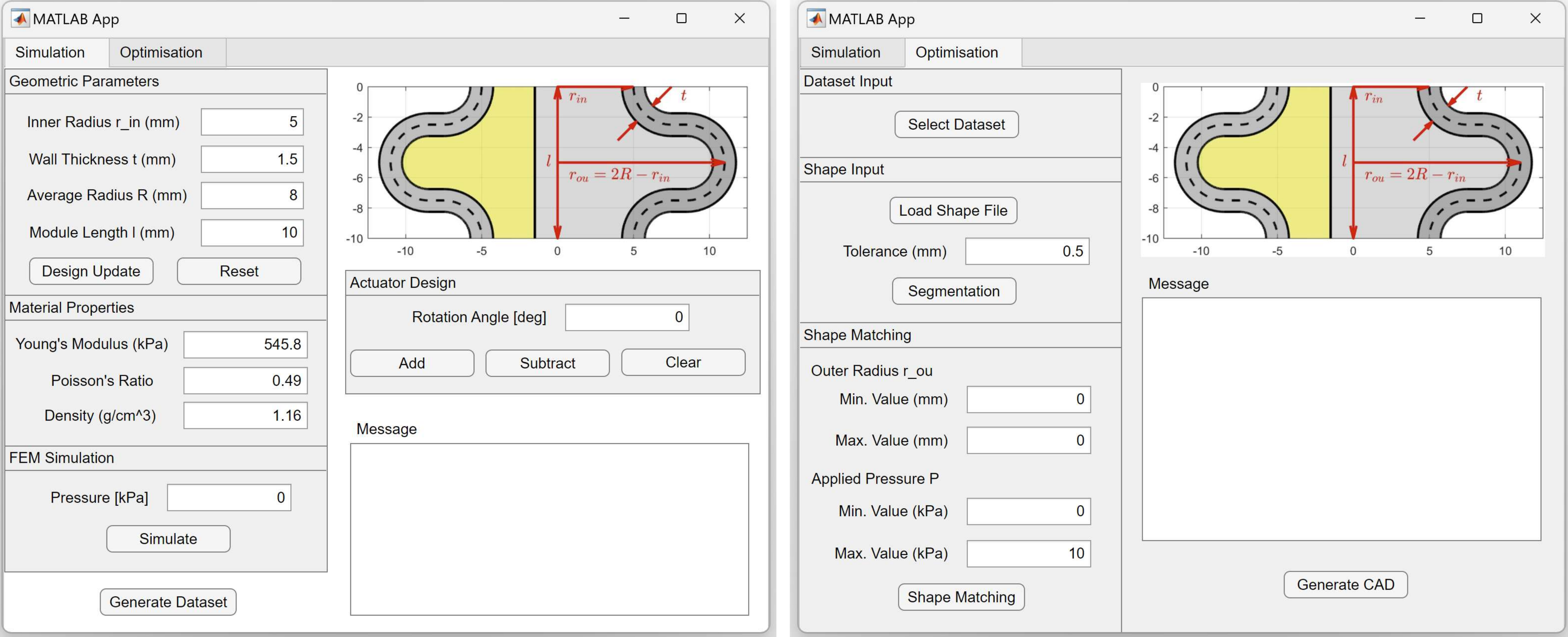}
    \caption{The graphic user interfaces of the toolbox SPADA.}
    \label{fig:GUI}
\end{figure*}

To amplify the efficiency and user accessibility of our design framework, we've developed SPADA --- an open-source design toolbox with a user-friendly GUI (see Fig.\ref{fig:GUI}) --- built on MATLAB\textsuperscript{\textregistered} and COMSOL MultiPhysics\textsuperscript{\textregistered}. Compared to our previously published bellow-SPA design toolbox --- designed solely for simulating bending actuator behaviors and taking approximately 74 minutes per simulation~\cite{Yao2022} --- SPADA stands out by facilitating simulations for diverse deformations in just a few minutes and offering efficient end-to-end shape-matching optimization. Here, “end-to-end” refers to the process from the desired shape to the STL files of the designed actuator for 3D printing.

The simulation component allows for the design of bellow SPAs by adjusting the modular geometric parameters, customizing assemblies, and modifying material properties. It utilizes background FEM simulation to analyze the behavior of modules and predict the actuator's configuration through forward kinematics. Additionally, it can generate material-specified datasets.

The optimization component can train a selected dataset into an ANN to serve as a surrogate model. It takes as input a file containing the desired shape of the actuator, represented as a series of ordered 3D points. The 3D PCC segmentation algorithm segments the desired shape into CC segments, and an optimization algorithm based on a genetic algorithm and the surrogate model is used to find the optimal actuator design parameters that approximate the desired shape. It can also generate CAD files of the designed actuator, ready for direct 3D printing.

See Supplementary Data C for detailed instructions on how to use SPADA. The source code is available on a GitHub repository~\cite{SPADA}, and a demonstration of the toolbox is provided in the Supplementary Video.

\section{Results}
\label{sec:results}

In this section, SPADA was used to design three actuators that accurately match the predefined 2D shapes with root-mean-squared-errors (RMSEs) of 4.16, 2.70, and 2.51mm, respectively, hence validating the accuracy of the kinematics model, illustrating the efficacy of the shape-matching algorithm, and demonstrating its ability in achieving end-to-end from desired shapes to designed actuators for direct 3D printing. Furthermore, we harnessed the toolbox's potential in the design of 3D deformable actuators, specifically by designing an actuator according to an elephant-trunk-inspired helical shape.

\subsection{2D Shape Matching: Letter Writing}
\begin{figure*}
    \centering
    \includegraphics[width=0.99\textwidth]{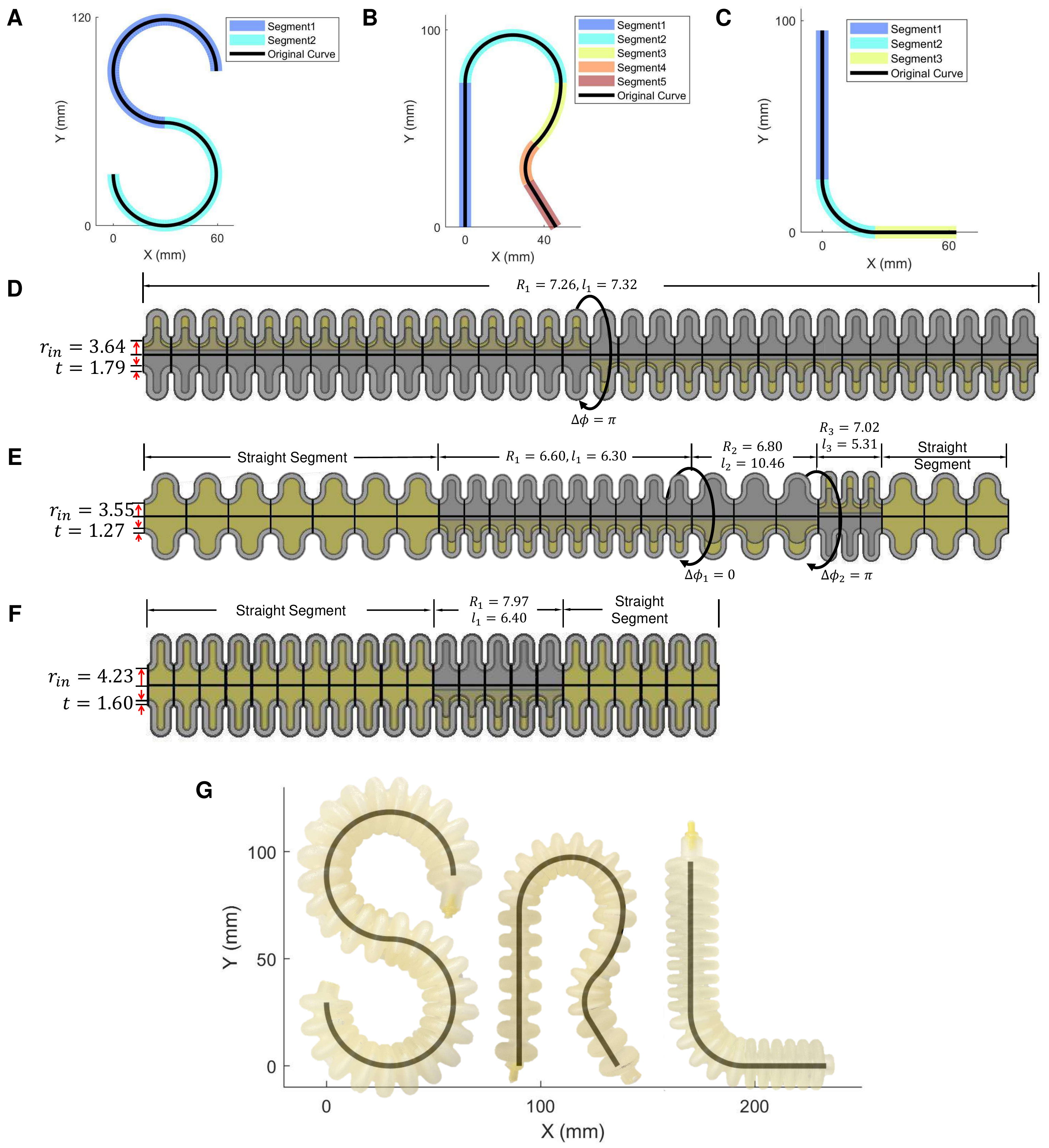}
    \caption{The 2D shape-matching case: the input shapes of letters "SRL" are segmented in A)-C). D)-F) are the actuator schematics that match the input shapes and are designed by optimal parameters. G) shows the deformed prototypes compared with the input shapes.}
    \label{fig:SRL}
\end{figure*}

In the 2D case of validating the shape-matching algorithm along with the kinematics model, and also demonstrating the ability of end-to-end from the desired shapes to 3D printed actuators, SPADA was used to design three actuators that match the shape of the letters "S", "R", and "L", which is the acronym of Soft Robotics Lab.

The shape for each letter is represented by a series of ordered 3D coordinates, imported into SPADA as a ".mat" file. The segmentation results for each letter are highlighted in Fig.~\ref{fig:SRL}A-C. The letter "S" is segmented into two arc segments with the same length and curvature. The letter "R" is segmented into two straight segments and three different arc segments. The letter "L" is segmented into two straight segments and one arc segment. All the lower bound values of the outer radius were set to be 9mm for three actuator optimization for ease of manufacturing. After optimization, the optimal design parameters are labeled on the schematics of actuators designed by those parameters in Fig.~\ref{fig:SRL}D-E. The surrogate model used was provided with the toolbox, that was trained by a provided FEM dataset of the default material Agilus30\texttrademark ~(see information of the FEM and material parameters in Supplementary Data A). The prototypes were 3D printed on a Stratasys J735\texttrademark ~using the CAD files generated by SPADA.

The shape of the actuated prototypes was compared to the initial desired shape in Fig.~\ref{fig:SRL}G by overlapping them in the x-y plane. The respective RMSEs between the desired shapes and actuators’ real shapes are 4.16, 2.70, and 2.51mm, validating and demonstrating our shape-matching algorithm (refer to Supplementary Data D.I for details of the experiments, and see the prototypes taking the defined shapes in the Supplementary Video available on the project's GitHub page~\cite{SPADA}).

\subsection{3D Shape Matching: Elephant Trunk-Inspired Shape}
The previous 2D case of shape-matching validates the efficacy of our end-to-end design framework, given its independence from gravitational effects. Yet, in the 3D scenario, gravity is unavoidable. Furthermore, gravity acts unidirectionally while the deformation of actuators significantly varies in space, making it exceedingly challenging to incorporate gravity's impact into the modeling process, particularly in modular approaches. This complexity explains why nearly all shape-matching design frameworks sidestep considering gravitational effects~\cite{Connolly2017, Singh2020, programmable3}. While Jiang et al.~\cite{Jiang2021}did incorporate gravity into their analytical model, they were only able to roughly approximate its influence for two types of segments and didn't evaluate it in the real world.

\begin{figure*}
    \centering
    \includegraphics[width=0.99\textwidth]{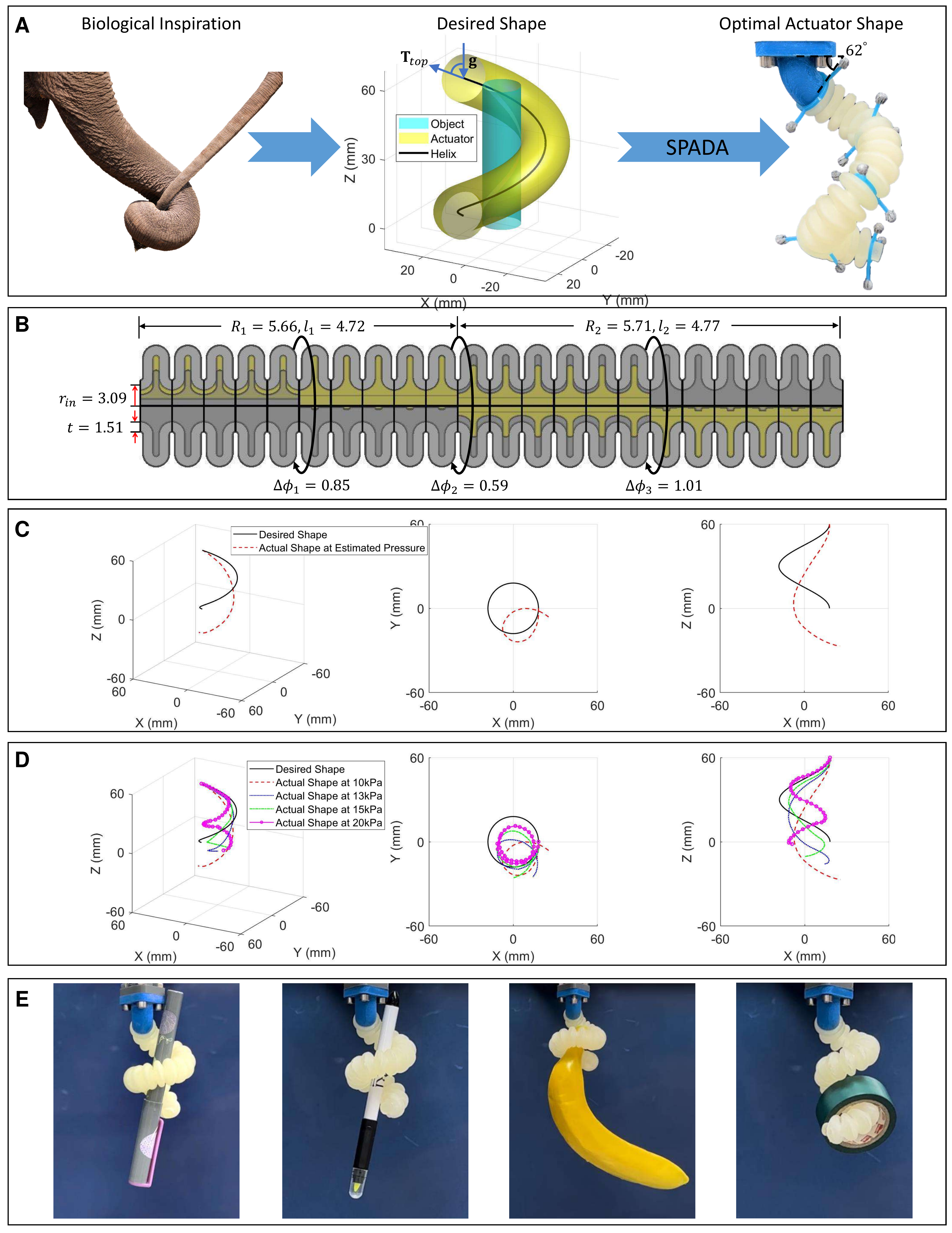}
    \caption{The 3D shape-matching case: A) depicts the process of inspiration drawing from an elephant trunk (edited image \copyright{} Shutterstock, accessed on August 1, 2023), acquiring the helical shape, and subsequent design of the actuator using our toolbox SPADA. B) showcases a schematic of the actuator with optimal design parameters. C) provides a comparison between the desired helical shape and the actual shape of the actuator under the pressure estimated by the toolbox. D) compares the desired shape and the actual ones of the actuator under escalated pressure. E) demonstrates the actuator's versatility in handling and grasping various objects.}
    \label{fig:trunk}
\end{figure*}

We recognize that shape-matching, on its own, can't offer a comprehensive solution for tasks involving gravity and contact. Yet, our design framework and toolbox still provide valuable insights for actuator design, even without considering these factors, thanks to the adaptability of soft actuators. The design of actuators to assume a helical shape, emulating the versatile and secure grasping behavior of an elephant's trunk, is a prime example among the research community~\cite{manipulation2, PneuNet3DPrint_ElephantTrunk, trunk}. Hence, to both demonstrate the potential of the toolbox for employing 3D shape-matching in guiding actuator design, and to quantitatively assess gravity's impact on the 3D shape-matching scenario, we delve into an end-to-end example of designing a helical actuator, inspired by the structure of an elephant's trunk, for the purpose of grasping irregular objects, as shown in Fig.~\ref{fig:trunk}.

A right-handed helix, with $x$, $y$, $z$ coordinates represented by the mathematical expression ($a\cos{t}$, $a\sin{t}$, $ht$) where $t$ is evenly spaced between 0 and $2\pi$, is utilized to embody the helical configuration inspired by the grasping movement of an elephant's trunk. By presuming that the object targeted for grasping fits within a cylindrical dimension of 16mm diameter and 120mm height, and further constraining the actuator's outer radius to be approximately 10mm, the helix parameters $a = 18$mm and $h = 60$mm can be derived for a power grasp.

The helical shape, consisting of a sequential of ordered 3D points, is imported into SPADA.  An actuator is then designed to match this shape using the toolbox framework while disregarding gravitational effects.  Fig.~\ref{fig:trunk}B showcases the design parameters of the actuator, and the toolbox estimates the actuation pressure to be 9.85kPa (which is approximated to 10kPa in experiments due to the hardware limitations). To assess the discrepancy between the desired shape and the actual shape of the pressurized actuator under the effect of gravity, we executed a series of experiments. The top end of the actuator is fixed to a connector with an angle of 62 degrees (Fig.~\ref{fig:trunk}A). This angle corresponds to the divergence between the tangent vector at the helix's top end and the gravity axis. 

We first pressurized the actuator to 10kPa, a value estimated by the toolbox as optimal for achieving the helical shape. Fig.~\ref{fig:trunk}C illustrates the deviation between the desired helical shape and the actual configuration of the actuator at this pressure. The root-mean-squared error is calculated to be 25.64mm. 

Subsequently, we escalated the actuation pressure to 13kPa, 15kPa, and 20kPa.  Fig.~\ref{fig:trunk}D compares the desired shape and the actuator's actual shapes under these increased pressure values. The calculated root-mean-square errors for these pressures were 19.08mm, 18.69mm, and 23.77mm, respectively. Moreover, as observed from the XY and XZ perspectives in Fig.~\ref{fig:trunk}C and D, it's evident that as pressure increases, the central point of the actual shape (XY view) approaches that of the targeted shape. Concurrently, the number of convolutions in the actual shape (XZ view) increases, while the height of the actual shape along the z-axis decreases. This phenomenon can be attributed to the escalating pressure that not only counteracts gravitational effects but also enhances the curvature perpendicular to the direction of gravity. This, in turn, induces an increase in the number of helical turns.

However, despite the error caused by gravity, it is still possible to exploit behaviors arising from the design and interaction with the object to achieve a successful grasp. The grasping experiments were performed to explore the potential of the designed actuator for handling various objects, as shown in Fig.\ref{fig:trunk}E. Refer to Supplementary Data D.II for details of the grasping experiments and watch the prototype grasping objects in the Supplementary Video.

\section{Conclusion}
In this work, an end-to-end design framework for the shape-matching of bellow-SPAs is proposed and implemented as an open-source design toolbox, SPADA. A model leverages FEM-simulated module deformation and PCC approximation to predict the kinematics of actuators. Surrogate modeling involving ANN training on a FEM dataset is applied to speed up computation. A 3D PCC segmentation algorithm approximates the desired curve by dividing it into CC segments.  An optimization algorithm, grounded on a genetic algorithm and a surrogate model, determines the optimal design parameters of the actuator to align with the shape of these segments. The toolbox, based on the proposed design framework, has proven its end-to-end capability in designing actuators to accurately match 2D shapes, while also demonstrating its potential to design deformable actuators in 3D space.

Overall, this design framework can be generalized to other soft actuators, such as tendon-driven soft actuators. Users can also specify more material properties for the actuator and use the toolbox FEM simulation function to create a data set for training. A GitHub repository has been created with the toolbox and locations for users to upload self-acquired datasets for co-design. In the future, the authors will maintain and update the toolbox with consideration of the gravity effect and environmental parameters (such as underwater conditions). Add-ons for protocols to design soft actuators for specific tasks will also be investigated, such as soft grippers, soft actuators for surgical operations, and soft crawlers for locomotion.

\newpage
\section*{Acknowledgements} 
The authors thank Dr Alessandro Albini (University of Oxford) for helping with the experimental setup and Mr Yuwen Chen (University of Oxford)  for advice on optimization algorithms.

\section*{Authorship Confirmation Statement}

Y.Y., L.H. and P.M. conceived the concept and wrote the article. Y.Y. performed the simulations and optimization, carried out experiments and interpreted the results. L.H. and P.M. directed the project. All authors commented on the article.

\section*{Author Disclosure Statement}

No competing financial interests exist.

\section*{Funding Statement}

This work was supported by the Engineering and Physical Sciences Research Council (EPSRC) Grant EP/V000748/1.


\bibliographystyle{vancouver}
\bibliography{references.bib}
\end{multicols}
\clearpage


\section*{Supplementary material}

# SUPPLEMENTARY DATA

August 21, 2023

## A  The FEM Simulation and Material Model

The FEM simulation is performed as a stationary study of the 3D solid mechanics model in COMSOL MultiPhysics® 5.6a. The half bellow module is modeled with the same parameters in Table. 1, with the incompressible Neo-Hookean hyperelastic material model in Eq. 3, factoring in uniaxial deformation.

The rationale behind the material model selection is twofold:

(1) It is a simple material model that only requires a shear modulus, which can be approximated by Young's modulus and Poisson's ratio, making it easy for users to find parameters for their own materials.

(2) This model offers a satisfactory agreement for lower strain values, typically below 50% as reported [25]. Furthermore, based on our previous investigations [31], the mechanical strain for bellow-SPAs primarily remains under 50% for an actuation pressure range of 0-10kPa. Consequently, the neo-Hookean model has been deemed an appropriate choice.

Regarding material properties, we've set Poisson's ratio to 0.49, and the material density $\rho = 1.16$ g/cm$^3$ is obtained from the official data sheet [24]. Unable to conduct direct experimental characterization of the material, we derived Young's modulus by linear fitting the Ogden model data of Agilus30™, as presented in the work of Abayazid and Ghajari [36], within a strain range of 0-1. This range was chosen as the data exhibited a consistent slope up to this point, with a noticeable change in slope for strains exceeding 1. As we couldn't access the original experimental data from their work, our comparison between the neo-Hookean model and their Ogden model for the given strain range yielded a Root Mean Square Error (RMSE) of 20% as indicated in Fig.S1 below.

Furthermore, the accuracy of the kinematic shape-matching is affected by the accuracy of the material model. From the experimental results of the 2D shape-matching example, it is possible to notice that despite the discrepancy between the two material models, we achieved a very good match.

1

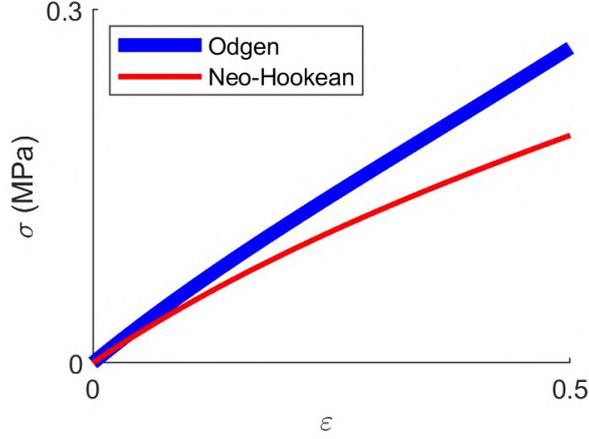

Figure S1: The comparison between the Ogden model data (from the work of Abayazid and Ghajari) and the Neo-Hookean model (using the linearly fitted Young's Modulus) for Agilus30™ in the strain range of 0-0.5.

In the FEM model, the cylinder used to smooth its element deformation has a radius of $r_{in} + \frac{t}{2}$ and a length of $l$ connected to the smaller end of the half bellow using the same material model. The pressure is modeled as the uniformly distributed load on the inner surface of the half bellow, ramped by intervals of 0.5 KPa. The mesh is set as the default element size of "Finer". To allow for large deformation of elements, the geometric nonlinearity is included, with "Constant Newton" and "Anderson acceleration" set as the nonlinear simulation method.

Assuming the plane where the cylinder is connected to the bellow is parallel to the plane at the other end of the cylinder, the angular deflection of the half bellow is the angle between this plane and the plane at the fixed end of the half bellow. Therefore, the angle can be calculated by the displacement of the top point $E_1$ and the bottom point $E_2$ in the plane at the free end of the cylinder (as shown in Fig. S2(A). If the coordinate of $E_1$, $E_2$ are $(x_1, y_1, z_1)$ and $(x_2, y_2, z_2)$, respectively. Then the angular deflection of the module can be calculated by

$$\theta = \arccos \frac{z_1 - z_2}{\sqrt{(z_1 - z_2)^2 + (x_1 - x_2)^2}} \tag{S1}$$

Additionally, the FEM simulated stress distribution, angular deflection and the strain distribution of the bellow SPA model with default design parameters in relation to the change of pressure are shown in Fig. S2.

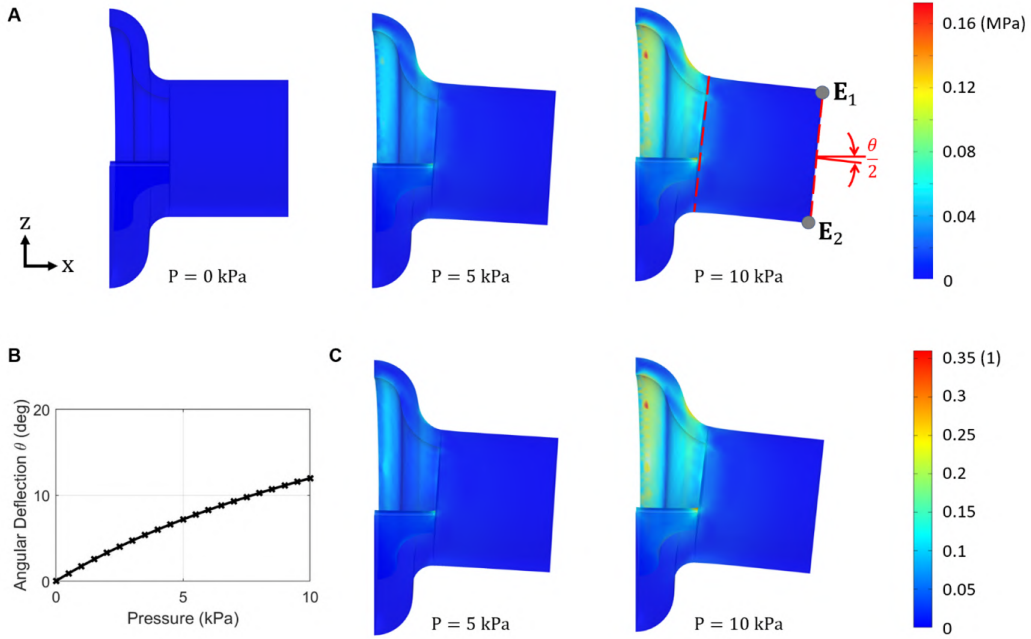

Figure S2: The detailed FEM simulation results of the bellow SPA module with the default geometric parameters (Table. 1) and the default material Agilus30™. (A) The von Mises stress distribution of half FEM model changes with increased pressure. (B) The module's angular deflection $\theta$ predicted by the FEM model changes with increased pressure. (C) The first principal strain distribution of half FEM model changes with increased pressure.

## B  The Optimization Algorithm

Numerous optimization algorithms are available, as detailed in the review [26]. Based on our prior investigation into optimization algorithms suitable for bellow-SPAs shape-matching design [41], we identified the genetic algorithm as a prime choice for this work. This choice was influenced by its compatibility with surrogate modeling and its efficiency in handling large design variable dimensions.

Despite being provided in our previous work, we revisit the optimization algorithm here by integrating a genetic algorithm, a surrogate model, and the symbols introduced in this study. Assuming there are in total $N$ unique arc segments with length and curvature pairs $[(L_1, \kappa_1), (L_2, \kappa_2), \ldots, (L_N, \kappa_N)]$ different from each other, there will be $N$ different bellow SPA modules, each approximating one segments.

We wish to determine their shared parameters $\mathbf{x} = [r_{in}, t, P] \in \mathbb{R}_+^3$ and a set of respective geometric parameters $\mathbf{y} = [\mathbf{y}_1, \mathbf{y}_2, \ldots, \mathbf{y}_N] = [(R_1, l_1), (R_2, l_2),$

3

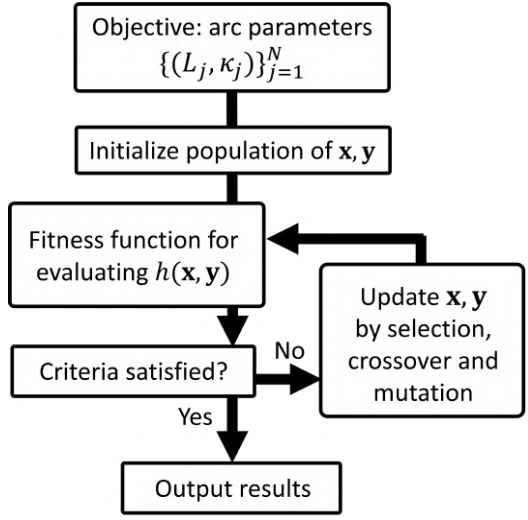

Figure S3: The optimization algorithm of designing bellow SPAs for shape-matching.

$\ldots, (R_N, l_N)] \in \mathbb{R}_+^{2 \times N}$. The optimization objective is to minimize discrepancies in arc parameters (arc length and curvature) between the desired shape $\{(L_j, \kappa_j)\}_{j=1}^N$ and the deformed shape of the actuator designed by $\mathbf{x}, \mathbf{y}$, which is constructed from designed modules and predicted by a surrogate model. We have a total of $2N+3$ optimization variables for one actuator, as shown in Fig. S3.

The constraints are specified by the maximum actuation pressure $P_{\max}$, the maximum curvature of segments $\kappa_{\max}$, the segment length and the inherent design constraints of the bellow SPA module. The problem is formulated as follows:

$$\min h(\mathbf{x}, \mathbf{y}) := \frac{1}{N} \sum_{j=1}^N d_j(\mathbf{x}, \mathbf{y}_j) \tag{S2}$$

$$\text{s.t.} \begin{cases} 0 \leq P \leq P_{\max} \\ 0 < r_{in} \leq \frac{1}{\kappa_{\max}} \\ \frac{R_j}{4} \leq t \leq \frac{r_{in}}{2} \\ r_{in} + t \leq R_j \leq \min(\frac{1}{2\kappa_j} + \frac{r_{in}}{2}, 2r_{in}) \\ 2t < l_j < \min(L_j, 4(R_j - r_{in})), \quad \forall j \in [N] \end{cases} \tag{S3}$$

where

$$d_j(\mathbf{x}, \mathbf{y}_j) = w_1 |L_j - n_j l'_j| + w_2 \left| \kappa_j - \frac{\theta_j}{l'_j} \right| \tag{S4}$$

in which $l'_j = \theta_j(\frac{l_j}{\theta_j} + r_{in} + \frac{t}{2})$ is module's deformed length, $n_j = \left[\frac{L_j}{l'_j}\right]$ is the

4

number of identical modules in the $j$-th segment, and $w_1 = \frac{1}{L_j}$, $w_2 = \frac{1}{\kappa_j}$ are the weighting factors used to balance the difference of units. $\theta_j = f(r_{in}, t, P, R_j, l_j)$ is obtained from the surrogate model trained before.

Beyond the default settings of genetic algorithms in MATLAB®, we established stopping criteria for either reaching 500 iterations or the objective function dropping below $0.02$. If a feasible point is found, the optimization outputs the shared parameters $[r_{in}, t, P]$ and the respective parameters $[R, l]$ for each segment.

# C  SPADA User Guide

## I  The Simulation Component

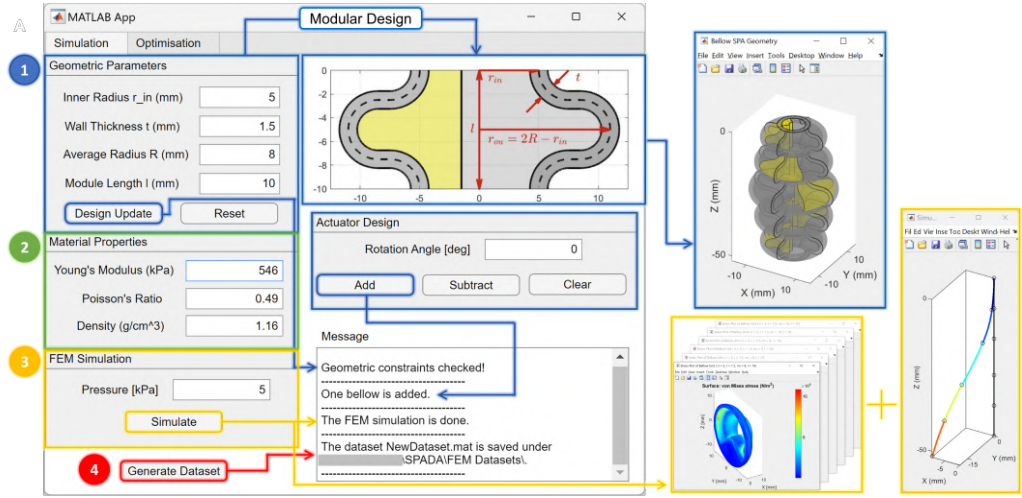

Figure S4: The GUI and usage instructions of the simulation component in SPADA (information for each step is highlighted in the same color as the step number icon)

The interface of the simulation component is shown in Fig. S4.

(1) The "Geometric Parameters" section allows entering four geometric parameters of the bellow SPA module. The "Design Update" button checks geometric constraints and updates the 2D schematic based on the input, while the "Reset" button resets them all to the default values. The "Actuator Design" section constructs the actuator in a modular approach (an example is shown in a blue box on the right side). The "Add" button stacks modules based on the current geometric parameters and orientation angles; "Subtract" removes the last added module, while "clear" removes all.

5

(2) The "Material Properties" section allows for defining the material of the actuator by its Young's modulus, Poisson's ratio and density.

(3) The "FEM Simulation" section takes the input of the actuation pressure; the "Simulation" button launches COMSOL MultiPhysics® in the background and analyzes the designed actuator using FEM. It simulates the behavior of each unique module and predicts the configuration of the whole actuator with forward kinematics (examples are shown in the first and second yellow boxes on the right side, respectively).

(4) The "Generate Dataset" button uses FEM to collect a dataset as explained in the paper's Section 2.B.I based on the input material properties. A dataset of Agilus30™ is already collected and provided to users.

## II   The Optimization Component

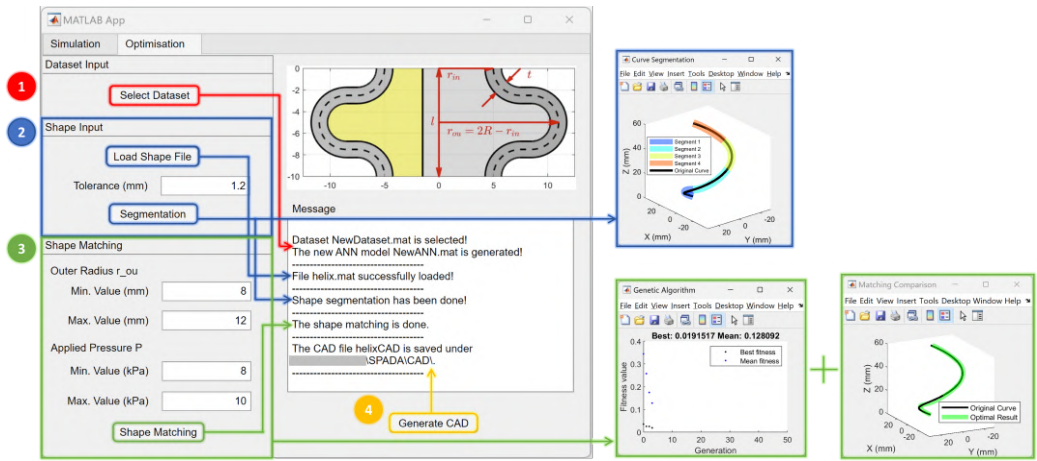

Figure S5: The GUI and usage instructions of the optimization component in SPADA (information for each step is highlighted in the same color as the step number icon)

The interface of the optimization component is shown in Fig. S5.

(1) In the "Dataset Input" section, the "Select Dataset" button trains a selected dataset into a surrogate model via an ANN as described in the paper's Section 2.B.II. The dataset can be obtained by executing step 4 in the toolbox's simulation component, with user-defined material properties. Alternatively, users can create their own datasets, ensuring matching the format of our provided dataset for Agilus30™ , available on our toolbox's GitHub repository [42].

(2) In the "Shape Input" section, the desired curve file, consisting of sequential 3D coordinates of points, can be input with the "Load Shape File" button. The "Segmentation" button helps divide the 3D shape into straight and circular arc segments by the 3D PCC segmentation algorithm according to the input tolerance value.

(3) The "Shape Matching" section uses a genetic algorithm to find and output the optimal design parameters of the actuator that matches the shape upon pressurization based on the input upper and lower bounds of the actuator's outer radius and the applied pressure.

(4) If the algorithm can find a set of feasible design variables, the "Generate CAD" button can generate a .stl format file of the designed actuator for direct 3D printing.

# D  Experimental Setup

## I  The 2D Shape-Matching

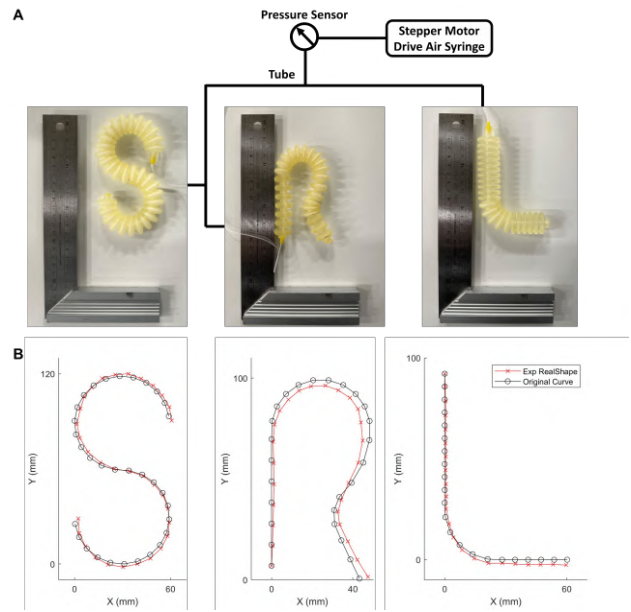

Figure S6: A) The setup for the 2D shape-matching experiment (top). B) The comparison between the original letter shapes and actuators' real shapes obtained through experiments (bottom).

7

The setup of the experiment in Sec 4.1 is shown in Fig.S6. Three actuators, "S", "R", "L" were driven with respective optimal pressures 8.11, 6.97 and 6.94kPa (which were approximated to 8, 7 and 7kPa in experiments due to the hardware limitations), determined along with their design parameters by the shape-matching design framework implemented on SPADA. A pneumatic system consisting of a 12V DC stepper motor driving a 60mL air syringe, a micro-controller (Arduino Uno), a pressure sensor (ADP51A11, pressure range 0-40 kPa, Panasonic, Japan) and a PI controller was used to control the input pressure. Lubricant was applied beneath the actuators to remove frictional effects. After taking photos from actuators' upward direction, the MATLAB Image Processing Toolbox™ was used to identify their real shapes by getting the central pixel coordinates of each module and using the steel square as reference. The original letter curves were sampled with the same intervals to compare with the real shapes. The root-mean-squared-errors for three actuators are 4.16, 2.70, and 2.51mm, respectively.

## II  The 3D Shape-Matching

The setup of the experiment in the paper's Section 4.B is shown in Fig. S7. The top end of the actuator is fixed to a connector (as shown in Fig.7A), which is connected to the top center of a 780×700×563 mm aluminum alloy frame. In order to get the deformed shape of the actuator, we 3D printed five rings, each featuring three evenly spaced rods connected to small reflective tape-coated spheres acting as markers for the OptiTrack™ system. Apart from the actuator's two ends, these rings are situated between every pair of adjacent segments. By identifying the position and orientation of these rings, and presuming a constant curvature between each pair, the real shape of the pressurized actuator can be obtained. The space coordinates of the markers were captured by a motion tracking system (OptiTrack™ with 4 Flex 3 cameras) and can be used to create a rigid body to represent the real-time positions of the free end.

A pneumatic system consisting of a 12V DC stepper motor driving a 60mL air syringe, a micro-controller (Arduino Uno), a pressure sensor (ADP51A11, pressure range 0-40 kPa, Panasonic, Japan) and a PI controller was also used to control the input pressure. We repeated the experiment 3 times for each trial. To analyze the data, we (1) calculated the average positions and orientations of all rings, which were treated as rigid bodies in the OptiTrack™ system, (2) determined arc planes and centers adjacent rings using their positions and tangent vectors, based on the constant curvature assumption, (3) aligned the position and orientation of the top ring with the uppermost point of the desired helical shape, (4) resampled the desired helical shape at the same interval as the sampled arcs' coordinates to derive RMSEs, comparing the desired and real shapes.

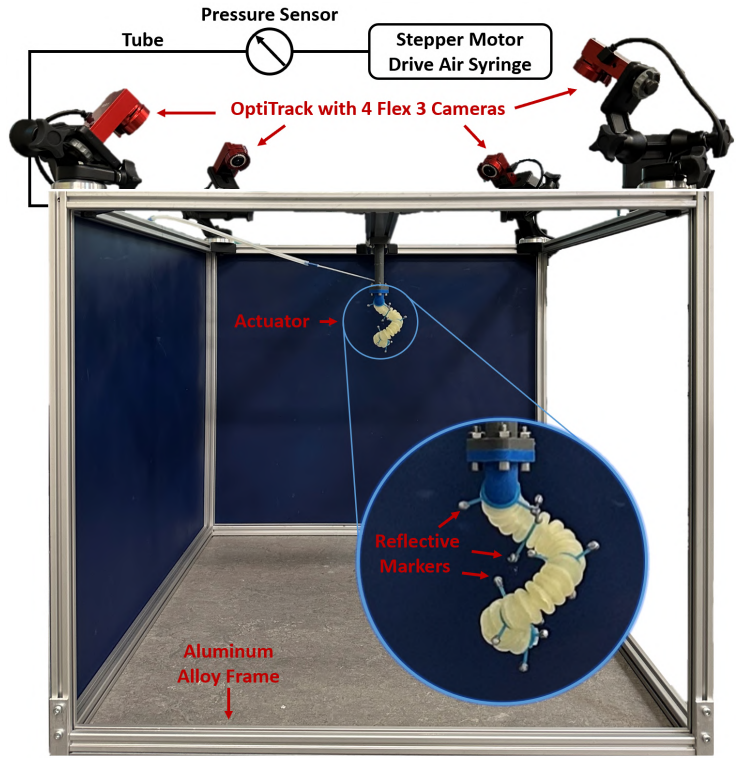

Figure S7: The experiment setup for the elephant-trunk-inspired, helical-shaped 3D deformable actuator.

To exploit grasping behaviors arising from the actuator design and interaction with the object, we performed grasping experiments on a series of objects. The object 1 in Fig.7E — a pen with a diameter of 16mm and length of 120mm, was the motivation for determining the helical shape's radius $a = \frac{16}{2} + r_{ou} = 8 + 10 = 18$mm, $h = 60$mm ensuring a power grasp. The experiment shows that the designed actuator can achieve a successful grasp for this object, which demonstrates the efficiency of our design pipeline. Subsequently, as shown in Fig. S8(A), we attached a 60mL syringe beneath the object and progressively increased its weight by filling it with water to explore the actuator's maximum grasping capacity. We discovered that when the syringe draws in water exceeding 30mL, equivalent to 73g (including the object), the grasp becomes unsuccessful. This is because grasping a heavier object requires higher pressure, which the designed actuator cannot handle for weights above 73g. Next, we evaluated the actuator's capability using another pen with smaller diameter shown in Fig.7E to assess its ability in grasping objects of reduced diameter. Following a successful grasp, we further tested its capability with even finer objects, confirming its ability to securely grasp items as slender as a zip tie shown in Fig. S8(B). Aside from power grasping, we also

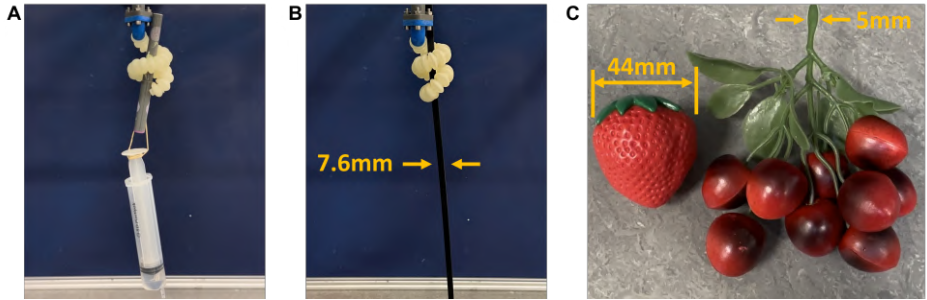

Figure S8: The supplemental objects for the grasping experiments: A) The object 1 in Fig. 7 carrying a syringe with water. B) A zip tie with a width of 7.6mm. C) Plastic strawberry with a diameter of 44mm and cherry with a handle of 5mm width.

explored the actuator's other grasping modes with object 3 and 4, as well as those shown in Fig. S8(C). Object 3 – a plastic banana represents a grasping mode where only a small part of the object, having a diameter within the preset range, is grasped. Despite the successful grasp of object 3, our attempt with a plastic cherry that has a slimmer handle, as seen in Fig. S8(C), was unsuccessful. This suggests that this grasping mode of the designed actuator has a diameter requirement for the object's contact region. Object 4 — a tape, represents another grasping mode where the object's geometry is not in the preset range as outlined in Fig.7A. However, the actuator, drawing inspiration from an elephant's trunk, coupled with the tape's mere 8mm thickness, could navigate through the tape's central hole and pinch it effectively. We attempted to use the pinch mode for grasping a plastic strawberry with diameter of 44mm but failed, indicating this grasping model of the designed actuator also has a requirement for objects' size.

In general, an actuator's grasping capability is influenced by its shape, material, and actuation pressure. Utilizing our toolbox necessitates a user-guided method to determine these aspects. By conducting grasping experiments and progressive refinements, users can achieve specific grasping performances. This iterative approach is also adaptable to various other applications.


\end{document}